\documentclass[sigconf]{acmart}
\usepackage{subfigure}
\usepackage{multirow}
\usepackage[ruled,linesnumbered]{algorithm2e}
\usepackage{graphicx}
\usepackage{array}
\usepackage{makecell}
\usepackage{caption}
\usepackage{hyperref}

\copyrightyear{2023}
\acmYear{2023}
\setcopyright{acmlicensed}\acmConference[CVMP '23]{European Conference on Visual Media Production}{November 30-December 1, 2023}{London, United Kingdom}
\acmBooktitle{European Conference on Visual Media Production (CVMP '23), November 30-December 1, 2023, London, United Kingdom}
\acmPrice{15.00}
\acmDOI{10.1145/3626495.3626497}
\acmISBN{979-8-4007-0426-0/23/11}

\acmSubmissionID{4}

\usepackage{ifthen}
\newboolean{final}
\setboolean{final}{false}

\begin{document}

\title{Expression-aware video inpainting for HMD removal in XR applications}


\author{Fatemeh Ghorbani Lohesara}
\email{ghorbani.lohesara@tu-berlin.de}
\affiliation{%
  \institution{Technische Universität Berlin}
   \country{Germany}
}

\author{Karen Eguiazarian}
\email{karen.eguiazarian@tuni.fi}
\affiliation{%
  \institution{Tampere University}
  \country{Finland}
}

\author{Sebastian Knorr}
\email{sebastian.knorr@eah-jena.de}
\affiliation{%
 \institution{Ernst-Abbe University of Applied Sciences Jena}
   \country{Germany}
  }

\renewcommand{\shortauthors}{F. Ghorbani, et al.}


\begin{abstract}
Head-mounted displays (HMDs) serve as indispensable devices for observing extended reality (XR) environments and virtual content. However, HMDs present an obstacle to external recording techniques as they block the upper face of the user. This limitation significantly affects social XR applications, specifically teleconferencing, where facial features and eye gaze information play a vital role in creating an immersive user experience. In this study, we propose a new network for expression-aware video inpainting for HMD removal (EVI-HRnet) based on generative adversarial networks (GANs). Our model effectively fills in missing information with regard to facial landmarks and a single occlusion-free reference image of the user. The framework and its components ensure the preservation of the user's identity across frames using the reference frame. To further improve the level of realism of the inpainted output, we introduce a novel facial expression recognition (FER) loss function for emotion preservation. 
Our results demonstrate the remarkable capability of the proposed framework to remove HMDs from facial videos while maintaining the subject's facial expression and identity. Moreover, the outputs exhibit temporal consistency along the inpainted frames. This lightweight framework presents a practical approach for HMD occlusion removal, with the potential to enhance various collaborative XR applications without the need for additional hardware.
\end{abstract}



\begin{teaserfigure}
  \includegraphics[width=\textwidth]{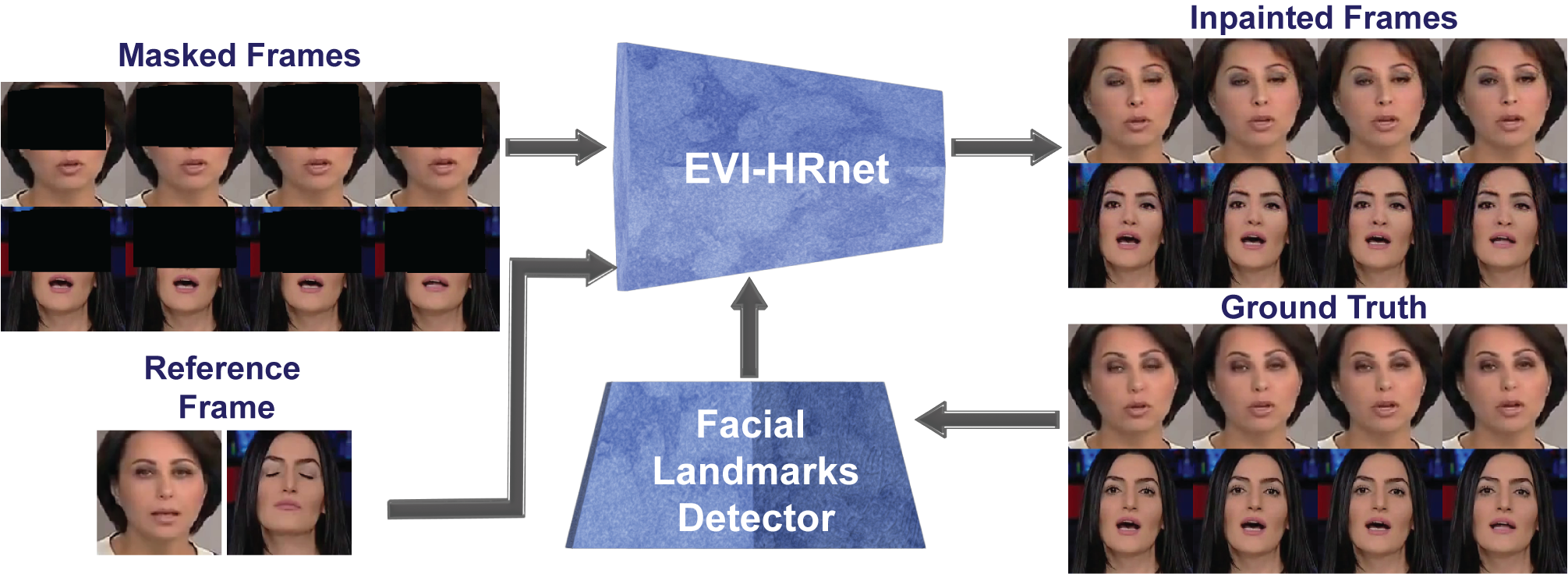}
          \caption{Overview of the pipeline of the proposed expression-aware video inpainting for HMD removal (EVI-HRnet) with the support of facial landmarks and a single occlusion-free reference frame. The examples depict the results from two distinct reference frames, considering scenarios where both eyes are either open or closed. Images: TSN (\url{https://www.youtube.com/watch?v=7sSPBQvxImQ}) and Al Aan TV (\url{https://www.youtube.com/watch?v=2iBnQpy8OMw}). 
  }
  \label{fig:pipeline}
\end{teaserfigure}

\maketitle

\section{Introduction}
\label{sec:intro}

The emergence of visual computing and immersive technologies has brought significant advancements in designing HMDs such as Meta Quest, Apple Vision Pro, Microsoft HoloLens, and HTC Vive. These HMDs have greatly enhanced the XR experience with the help of various sensors, including eye and gaze tracking. They aim to offer users a remarkable immersive and high-quality virtual environment interaction. However, HMDs can significantly hinder the virtual experience in social XR applications as the headset covers the person's upper face part, including eyebrows and eyes. 

Considering the diverse range of XR applications, especially for social and remote interactions such as virtual reality (VR) teleconferencing ~\cite{9848469}, collaborative education ~\cite{rojas2023systematic}, healthcare ~\cite{ali2023systematic}, and entertainment ~\cite{simons2023intelligence}, the removal of HMDs occlusion from multimedia content plays a crucial role in elevating the virtual communication experience for end-users and allowing eye contact. HMD removal describes the task of recovering the missing content of the faces, caused by the occlusion of an HMD, in a realistic way to enhance virtual communication.

While various approaches exist for representing humans in VR ~\cite{van2022deep, prins2018togethervr}, our research focuses on addressing the challenge of HMD removal to enrich the visual quality of human facial expressions representations in XR applications, which is the main objective of our paper. 
Based on the literature ~\cite{numan2021generative}, two categories of HMD removal approaches can be identified: model‐based approaches ~\cite{lou2019realistic, chen20223d, thies2016facevr}, which generate person-specific cartoon‐like avatars of human faces, and image‐based methods ~\cite{numan2021generative, wang2019faithful, zhao2018identity}, which result in nearly photorealistic outputs. 

In this paper, we focus on image-based methods and introduce a novel video inpainting pipeline based on the facial expressions in each frame to recover occluded face parts as depicted in Figure \ref{fig:pipeline}. While closely related to image inpainting, video inpainting of human faces is a task with many unsolved challenges such as proper recovery of the facial expressions and temporal consistency in videos. The existing image-based HMD removal approaches do not handle video frames properly as they lack modeling temporal consistent facial expressions \cite{numan2021generative, wang2019faithful, zhao2018identity}. 
To solve these problems, our pipeline is based on the recent progress in the field of video inpainting and uses a generative adversarial network (GAN) to inpaint occluded regions of faces. The inputs of the pipeline are masked frames (simulating the HMD), a single reference frame without masks, and the ground truth frames without masks. From the latter, the facial landmarks are detected and fed together with the masked frames and the reference frame into the EVI-HRnet as shown in Figure \ref{fig:pipeline}. Please note, that in a real application, no ground truth frames are available and that the facial landmark detector can be applied on masked frames to detect all visible landmarks of the unoccluded face parts and the occluded landmarks from frames captured inside an HMD with internal cameras. 
We compare our framework with LGTSM ~\cite{chang2019learnable}(our baseline model) and CombCN ~\cite{wang2019video}) and prove the model performance in inpainting large HMD occlusion across all the frames by using a publicly available facial video dataset ~\cite{rossler2018faceforensics}. Furthermore, we provide an ablation study to evaluate the influence of landmarks and our novel FER loss function for emotion preservation. To the best of our knowledge, EVI-HRnet stands out as the pioneer approach capable of restoring intricate facial details in videos, such as eye blinking, for the HMD removal task.  

The main contributions of this work can be summarized as follows: 
\begin{itemize}
\item We propose a GAN-based framework capable of inpainting time-consistent facial information with the help of a single unmasked reference image of the user to preserve his/her identity.  
\item We introduce a new subtle FER loss to further enhance the quality of the inpainted frames. 
\item We utilize the facial landmarks of each frame to handle distinctive facial features, such as the shape of the eyes and forehead, thus helping the model to generate more realistic and accurate outputs. 
\item By incorporating an attention module into the generator architecture of the proposed inpainting algorithm, the network
gains the ability to capture non-local features and long-range dependencies. This enhancement improves the performance of the network in
the task of HMD removal where a comprehensive understanding of the global context and unoccluded facial features is crucial for success.
\end{itemize}
The remainder of the paper is structured as follows: We first review the related work in HMD removal and video inpainting methods in Section \ref{sec:related_work}. Section \ref{sec:Tech_back} summarizes the fundamental concepts used for the design of our presented pipeline. Then, in Section \ref{sec:approach}, we describe the proposed method in detail. In Section \ref{sec:results}, we present and discuss the experimental results qualitatively and quantitatively together with an ablation study. Finally, the article is concluded in Section \ref{sec:conclusion} with a summary and future works.
\section{Related Work}
\label{sec:related_work}

In this section, we review the related work in HMD removal approaches, which is the application of our proposed method. 
As our approach is based on video inpainting, we broaden our review of the state-of-the-art and also include recent video inpainting methods.

\subsection{Head-Mounted Display (HMD) Removal}

The concept of HMD removal addresses the task of restoring the lack of image information caused by the occlusion of an HMD in a meaningful and lifelike manner ~\cite{numan2021generative}. Two distinct approaches for HMD removal exist: model-based and image-based methods~\cite{numan2020generative}. 

\subsubsection{Model-based face reconstruction}

In ~\cite{lou2019realistic}, the authors first reconstruct a fully textured 3D face from a user’s image. A classifier is learned for assessing emotions from six action units. Electromyography (EMG) sensors are attached to the HMD for tracking facial muscle movements and expressions. However, EMG data is not highly accurate for expression recognition. In another study ~\cite{chen20223d}, the authors introduced a person-specific real-time system that is able to reconstruct 3D faces with HMDs, and robustly capture eye gaze. They fitted the display with three infrared (IR) cameras. Two internal cameras for capturing the left and right eye images, and one external camera for capturing the unoccluded facial motion. The external camera is attached to the HMDs, as well as cell phone sensors to track head rotations. Five infrared LED lamps are also placed internally to each side of the device to provide consistent brightness.

The FaceVR system ~\cite{thies2016facevr} reconstructs the face model when the user is wearing an HMD. The authors placed two infrared cameras inside the HMD to capture eyes and one RGB-D camera outside the HMD to record the complete face of the person. Similar to this work, a capturing setup is presented in ~\cite{chen2018real} to acquire the facial features in order to create an avatar-based model of the user. This approach also needs infrared cameras fixed inside the HMD to properly reconstruct the user’s identity and eye movement, and it requires calibration for each person ~\cite{10.1145/3084363.3085083}.

\subsubsection{Image-based face completion}

Recent works ~\cite{chen2023multi,chen2023dgca,lahiri2020prior,yu2019free} have adopted the concept of GANs ~\cite{goodfellow2014generative} for image completion, which learns a representative estimate of the distribution of the given training data. Moreover, they also have been employed to solve the HMD removal problem ~\cite{gupta2022attention, gupta2022supervision}. In ~\cite{numan2021generative}, the authors introduced a framework based on GANs, which is capable of RGB-D face image inpainting for HMD removal. Throughout the model training process, the discriminator learns to determine inpainted images from ground truth images, and it takes a masked RGB-D image and a binary mask as its input. The discriminator of this architecture provides a means of focusing on different locations and semantics across image channels. The coarse-to-fine network, which acts as the generator, tries to inpaint the RGB-D image in two steps. The first step delivers a coarse prediction of the masked image region. This prediction is then fed to the second step where it is further refined. They used a customized synthesized data set due to the lack of suitable RGB-D datasets.  However, this framework does not take temporal consistency into consideration and is not directly applicable to video frames. 

In a similar study based on GANs for RGB image inpainting ~\cite{wang2019faithful}, the authors introduced a framework that uses a facial landmark detector. The detected facial landmarks are then passed to a GAN architecture, combined with an occluded RGB image and a reference image. To fill the occluded region in the face image, they first estimate the head pose features from the input image and the reference image, respectively. After that, they provide the input and reference images for a completion network conditioned on the assessed head pose attributes to render an image with the filled missing region. Nevertheless, as the authors mentioned, their method separately takes each frame for inpainting without considering temporal consistency, i.e. jittering and flicking artifacts can be noticed for video applications.

Moreover, in ~\cite{zhao2018identity}, the authors proposed a GAN-based architecture that is able to preserve the subject’s identity after HMD removal using a reference image of the target subject for face completion. They regularize the generator using two discriminators: a global discriminator that executes context consistency between filled pixels with background details, and a pose discriminator that regularizes the high-level postural errors. The pose map is the input of both the generator and the condition of the pose discriminator. As the authors described, jittering can be seen in the transition between frames in the video inpainting results because the temporal consistency is not explicitly constrained in their formulation, thus leaving it as a future work. In our work, we particularly pay attention to expression-aware and time-consistent video inpainting in faces.

\subsection{Video inpainting}

Video inpainting can be considered as an extension of image inpainting, incorporating temporal constraints to ensure consistency across different frames ~\cite{yang2023deep, szeto2022devil, chang2019free, chang2019learnable, wu2020image, wang2019video}. Despite the extensive research conducted on image inpainting, video inpainting poses additional challenges that yet need to be fully resolved. Furthermore, the majority of existing studies have focused primarily on object removal and scene inpainting ~\cite{zou2021progressive}, neglecting the specific domain of facial video inpainting involving human subjects, which presents its own set of complexities due to the intricate nature of facial features and the familiarity of faces making a plausible completion more challenging.

Moreover, despite the overall efforts in the field of video inpainting and their promising outcomes, they have mainly revolved around the removal of moving objects or people, i.e. moving masks, across frames, resulting in changing the location of the occluded regions throughout the video sequence. However, in the context of HMD removal, the occluded region remains unchanged in a virtual application, such as teleconferencing, since, when the user wears an HMD, the upper part of the face is continuously occluded. Consequently, the current video inpainting solutions do not address the problem of HMD removal and necessitate adjustments to be practical for XR applications.       
\section{Technical background}
\label{sec:Tech_back}
In this section, we describe the technical background of representing temporal features in our approach based on the current solutions. We then present the necessity of facial landmarks for HMD removal and our direction for acquiring them.  

\subsection{Temporal modeling}

For modeling temporal features, we employ the Temporal Shift Module (TSM) ~\cite{lin2019tsm}, initially developed for action recognition, and it has been further used for video inpainting. TSM tackles temporal knowledge in 2D convolutions with less complexity but can gain the performance of 3D convolution neural networks (CNN). For each frame, TSM moves a portion of feature channels to its earlier and subsequent frames before the convolution procedures. These shifted channels include attributes from neighboring frames, i.e., together with unshifted features. Thus, the original 2D convolutions can learn the temporal details consequently.

In ~\cite{lin2019tsm}, the authors describe the TSM operation with normal 1-D convolution with a kernel size of 3 for simplicity. Assuming the convolution weights are denoted by $W=(w_1, w_2, w_3)$, and the input $X$ is a 1-D vector with an unlimited length. The convolution operator $Y=\text{Conv}(W, X)$ can be mathematically represented as follows:
\begin{equation}
Y_i = w_1 X_{i-1} + w_2 X_{i} + w_3 X_{i+1}.
\end{equation}
We can decompose the convolution operation into two distinct steps: shift and multiply-accumulate. In the shift step, we displace the input $X$ by $-1, 0, +1$, and then proceed to multiply by $w_1, w_2, w_3$ respectively. These operations collectively yield the output $Y$. Precisely, the shift operation can be described as follows:
\begin{equation}
\vspace{-5pt}
X^{-1}_i = X_{i-1},~~~~ X^{0}_i = X_i, ~~~~ X^{+1}_i = X_{i+1}.
\end{equation}
Concurrently, the multiply-accumulate step is described by the equation:
\begin{equation}
\vspace{-5pt}
Y=w_1 X^{-1} + w_2 X^{0} + w_3 X^{+1}.
\end{equation}
The initial shift phase can be executed without any need for multiplicative operations. On the other hand, the subsequent multiply-accumulate phase is computationally more intensive. However, TSM conveniently integrates the multiply-accumulate process into a subsequent 2D convolution. As a result, this integration incurs no additional computational overhead when compared to 2D CNN-based models.

TSM efficiently helps both offline and online video inpainting. In our scenario for HMD removal, we only shift from past frames to current frames in a uni-directional manner (online TSM) since HMD removal in social XR applications should eventually be real-time capable.

\subsection{Facial landmark detection}
\label{sub:FLD}

To address the HMD removal problem, the utilization of facial landmarks proves essential. These landmarks offer crucial information about the facial structure, aiding in the accurate recovery of facial expressions and other facial attributes covered by the HMD. To recover facial expressions with EVI-HRnet, we follow a series of steps. Firstly, we employ an MTCNN (Multi-task Cascaded Convolutional Networks) ~\cite{zhang2016joint} face detector to detect faces within the input data. This detector accurately identifies the locations of faces present in the image.

Next, we utilize a facial landmark predictor to detect and locate specific facial landmarks within the detected faces ~\cite{dlib09,kazemi2014one}. The predictor employs a predefined model that is trained to identify 68 distinct points on the face, corresponding to key features such as the eyes, nose, and mouth. By mapping the detected facial landmark regions to their associated landmarks according to the 68-point model, we obtain precise information about the facial structure and geometry.

Finally, leveraging the shape of the input image and the set of obtained facial landmarks, we connect the landmarks with lines. This process effectively creates a contour that represents the facial structure. By connecting these landmarks, we can accurately depict the facial expressions and capture the nuances of the person's face, helping the generator to further recover the missing information.

It is worth noting that while our method employs a certain facial landmark detection approach, it is not the only available option. More accurate facial landmark detectors can also be integrated into our framework, further enhancing the precision and effectiveness of the facial expression recovery process.
\section{Approach}
\label{sec:approach}
\begin{figure*}[t]
    \centering
    \includegraphics[width=0.85\linewidth]{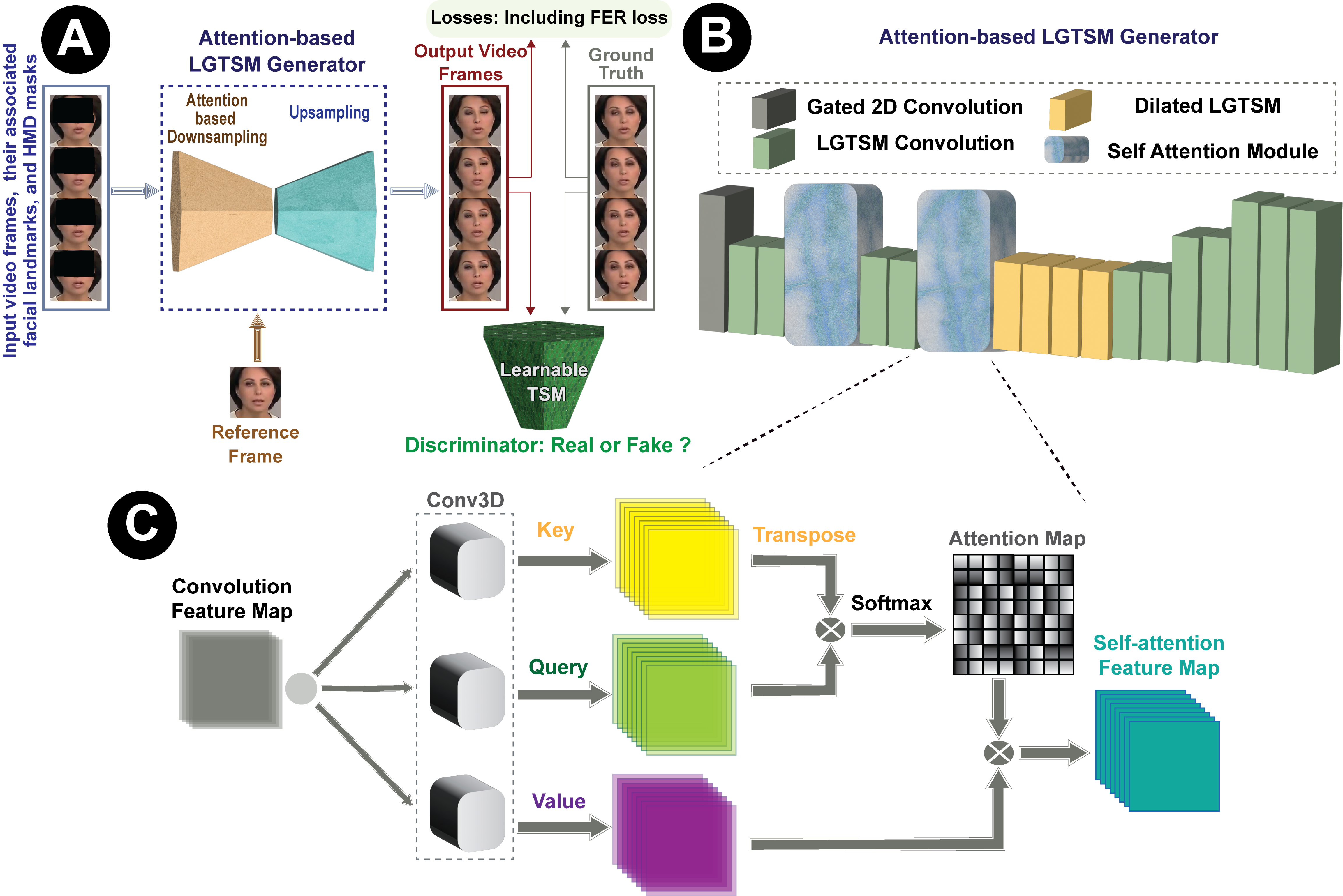}
    \caption{The proposed framework of EVI-HRnet: 
    (A) The overall inpainting network including a generator and a discriminator, (B) details of the attention-based LGTSM generator architecture, and (C) the components of the attention module (images: TSN (\url{https://www.youtube.com/watch?v=7sSPBQvxImQ})).}
    \label{fig:architecture}
\end{figure*}
\subsection{Framework architecture}
We utilize the Learnable Gated Temporal Shift Module (LGTSM) model for video inpainting proposed by ~\cite{chang2019learnable} as our base architecture. Our decision is based on its state-of-the-art performance without requiring additional parameters from 3D convolutions, optical flow information, and its architectural characteristics. LGTSM is employed to let 2D convolutions efficiently take advantage of neighboring frames, which is vital for video inpainting. Particularly, in each convolution layer, LGTSM learns to shift some input channels of the convolutional layers to its temporal neighbors, i.e. 2D convolutions can be improved to address temporal features. A gated convolution is also operated in the layer to determine the masked regions that are poisoning conventional convolutions. 

However, information aggregation of LGTSM only functions sufficiently in the local spatial–temporal region but fails to aggregate non-local information due to the bias of convolution. To solve this problem, we include an attention mechanism in the model to handle non-local features as introduced in ~\cite{zhang2019self}. Attention mechanisms allow the network to focus on different parts of the input data when making predictions. By incorporating this mechanism, the network gains the ability to capture non-local features and long-range dependencies within the feature maps, improving its performance in tasks that need an understanding of the global context. 

EVI-HRnet, shown in Figure ~\ref{fig:architecture}, follows the design of the base architecture ~\cite{chang2019learnable}. The proposed architecture includes an attention-based LGTSM generator and a TSM patch discriminator, which are trained via an adversarial process. In this process, the goal of the generator is to correctly inpaint a masked frame using a single unoccluded reference image and facial landmarks, and the aim of the discriminator is to specify whether the final inpainted frame is real or fake. By incorporating a facial expression recognition loss during training, the generator can further improve the quality of the inpainted frames. In the following, we go through each component of the model and explain the model pipeline in more detail. 

\subsection{Input data}

The inputs of the pipeline are masked frames (simulating the HMD), a single reference frame without masks, and the ground truth frames without masks. From the latter, the facial landmarks are detected and fed together with the masked frames and the reference frame into the EVI-HRnet, as we described in Section ~\ref{sub:FLD}. The incorporation of an RGB reference face frame into the generator's input is to address the HMD removal problem while ensuring the preservation of a person's identity. This reference image plays an important role in enabling the generator to accurately fill in the missing or occluded area, taking into account the individual's identity.

We choose the first frame in the sequence as the single reference image to establish the reference frame. Thus, we require that a user takes an image of his face without wearing an HMD before starting e.g. a social XR application, such as a VR teleconferencing, which is a valid practical requirement. However, it is worth noting that the reference frame can also be selected randomly from among the available frames. This flexibility allows for variations in the reference frame selection process, while still maintaining the necessary input for the generator to perform its inpainting task accurately. 

As we only need the occluded part (HMD occlusion) in the reference frame for inpainting, we zero out the pixels in the reference frame that are outside the HMD mask area.
\subsection{Generator}

The generator is composed of 13 convolution layers with the gated TSM, consisting of attention-based down-sampling, dilation, and up-sampling. We use the self-attention layers after the first down-sampling and before dilation, respectively, as depicted in Figure \ref{fig:architecture}(B). The attention weights are computed by comparing the similarity between spatial locations using key, query, and value transformations as shown in Figure \ref{fig:architecture}(C). The key, query, and value transformations are linear projections applied to the input feature maps using 3D convolution. The key and query transformations capture spatial relationships and dependencies, while the value transformation encodes the feature information. 
We compute query ($Q$), key ($K$), and value ($V$) representations from the input feature maps as follows:
\begin{equation}    
    Q = X \cdot W_Q,~~~~ \quad K = X \cdot W_K,~~~~ \quad V = X \cdot W_V.
\end{equation}
where $W_Q$, $W_K$, and $W_V$ are learnable weight matrices.
We then calculate the attention scores ($S$) by computing the dot product between the queries and keys, and apply softmax along the channel dimension (with $axis=2$ resulting in the attention score map of the same shape as the input feature maps) to obtain normalized attention scores ($A$):
\begin{equation}    
     S = Q \cdot K^T,~~~~ A = \text{softmax}(S, \text{axis}=2).
\end{equation}
These weights are finally employed to linearly combine the value maps, resulting in the attention feature maps ($Y_A$):
\begin{equation}    
     Y_A = A \cdot V.
\end{equation}

The attention feature maps are then passed through the subsequent layers of the network for further processing ~\cite{zhang2019self}.
There is no skip connection in the down-sampling layers as there are various masked regions, and we need to keep the features. For down-sampling and up-sampling layers, bi-linear interpolation before the convolutions is applied. In the whole training process, the generator G takes an occluded RGB frame, a reference frame, facial landmarks, and a binary mask of the occluded region as its input and outputs the inpainted frames.

\subsection{Discriminator}

During the learning process, the discriminator learns to determine inpainted frames from ground truth frames. The discriminator takes an occluded RGB frame and a binary mask as its inputs in order to learn the structure of the real data and the missing parts. The discriminator then encourages the generator to generate meaningful information in the occluded areas through an adversarial learning process.

We use the TSMGAN discriminator of the base model ~\cite{chang2019learnable}, that is composed of six 2D convolution layers with TSM.

\subsection{Loss functions}

We employ a combination of various loss functions to effectively train EVI-HRnet and achieve accurate results. The losses used in the model can be summarized as follows:

\begin{itemize}
    \item Reconstruction Loss (ReconLoss): This loss measures the reconstruction error between the model's outputs and the ground truth frames. It uses the L1 loss (mean absolute difference) between the output and ground truth frames.
    \item VGG Loss (VGGLoss): This loss calculates the perceptual difference between the output and ground truth frames based on the VGG network's feature maps pre-trained on ImageNet ~\cite{simonyan2014very,russakovsky2015imagenet}. It computes the L1 loss between the intermediate feature maps of the VGG network for both the output and ground truth frames.
    \item Style Loss (StyleLoss): This loss measures the style difference between the output and ground truth frames. It computes the L1 loss between the intermediate feature maps of the VGG network ~\cite{gatys2015neural} for both the output and ground truth frames.
    \item Adversarial Loss (AdvLoss): This loss calculates WGAN (Wasserstein GAN) loss which leverages the Wasserstein distance to create a value function being optimized by the model ~\cite{arjovsky2017wasserstein}. It computes the mean of the element-wise product of outputs and labels during the training of the discriminator.  
    \item Facial Expression Recognition Loss (FERLoss): We introduce a novel loss function, called FER loss, to train the model to accurately learn facial expressions. This loss evaluates the model's performance based on facial expression recognition. It computes the mean absolute error (L1 loss) between the predicted emotion scores of the inputs and outputs using an emotion recognition model.
\end{itemize}

Each loss is associated with a weight or scaling factor that determines its relative importance during training. These losses collectively contribute to the training of the model to generate visually appealing and emotionally expressive video outputs.

More in detail, for FER loss, we calculate the FER score for eight facial expression classes (namely surprise, angry, sad, contempt, disgust, fear, neutral, and happy) using an EfficientNet model ~\cite{savchenko2022video} trained on the AFEW (Acted Facial Expressions in the Wild) dataset ~\cite{dhall2012collecting}. This FER score serves as a measure of the predicted emotions in both the ground truth and output images (inpainted frames), allowing us to optimize the network during the training process. 

We further compute the L1 score, which represents the difference between the predicted emotions in the input and output images. This score assists in guiding the training process and improving the model's ability to accurately learn and replicate facial expressions. By integrating these various loss functions, we aim to enhance the performance of our model in terms of accurately inpainting facial expressions and producing realistic and emotionally consistent results.
In summary, the FER loss can be calculated for N number of samples in the batch as follows:
\begin{equation}
    \text{FER Loss} = \frac{1}{N} \sum_{i=1}^{N} \left| \text{$S_i$} - \text{$S_o$} \right|
\end{equation}
where $S_i$ are the FER scores in the input frames, and $S_o$ are the FER scores in the output frames generated by the inpainting network.
The overall loss function for training the model can be expressed as:
\begin{equation}
\begin{split}
    \text{Total Loss} = & \lambda_{\text{AdvLoss}} \cdot \text{AdvLoss} + \lambda_{\text{StyleLoss}} \cdot \text{StyleLoss} \\
    & + \lambda_{\text{VGGLoss}} \cdot \text{VGGLoss} + \lambda_{\text{FERLoss}} \cdot \text{FERLoss} \\
    & + \lambda_{\text{ReconLoss}} \cdot \text{ReconLoss}
\end{split}
\label{equ:loss}
\end{equation}
where $\lambda_{\text{AdvLoss}}$, $ \lambda_{\text{FERLoss}}$, $ \lambda_{\text{StyleLoss}}$, $ \lambda_{\text{VGGLoss}}$, and $ \lambda_{\text{ReconLoss}}$ are the weights for AdvLoss, FERLoss, StyleLoss, VGGLoss, and ReconLoss, respectively. 
\section{Experimental Results}
\label{sec:results}

\begin{table*}[!t]
  \centering
  \caption{Quantitative results of FaceForensics validation set with HMD masks. The metrics are averaged resulted from our model (EVI-HRnet), EVI-HRnet without the facial landmark usage, EVI-HRnet without inserting FER loss, the baseline model (LGTSM), and the CombCN model. {$^*$} These models are originally designed for inpainting moving masks across frames.}
  \resizebox{\textwidth}{!}{%
  \begin{tabular}{p{2cm} p{2cm} p{2.5cm} p{2.5cm} p{3cm} p{3cm}}
    \hline
    \toprule
    \textbf{Model} & \textbf{EVI-HRnet} & \textbf{EVI-HRnet} & \textbf{EVI-HRnet} & \textbf{LGTSM{$^*$}} & \textbf{CombCN {$^*$}}\\ 
     &  & \textbf{w/o landmarks} & \textbf{w/o FER loss} & \textbf{\cite{chang2019learnable}} & \textbf{\cite{wang2019video}}\\
     \hline
    \textbf{MSE\(\downarrow\)} & \textbf{0.003} & 0.004 & 0.0045 & 0.0046 & 0.0062 \\ \hline
    \textbf{PSNR\(\uparrow\)} & \textbf{25.83} & 24.45 & 23.93 & 23.73 & 22.51 \\ \hline
    \textbf{SSIM\(\uparrow\)} & \textbf{0.8986} & 0.8708 & 0.8598 & 0.8602 & 0.8460 \\ \hline  
    \textbf{LPIPS\(\downarrow\)} & \textbf{0.0508} & 0.0658 & 0.0661 & 0.0847 & 0.1866 \\ \hline 
    \textbf{FID\(\downarrow\)} & \textbf{0.7675} & 0.8345 & 0.8709 & 0.8340 & 1.1684  \\ 
    \bottomrule
  \end{tabular}
  }
  
  \label{tab:metrics}
\end{table*}
We conduct two types of experiments to demonstrate the performance of our proposed network, EVI-HRnet. First, we prove the impact of the landmarks and the FER loss in EVI-HRnet performance with an ablation study. Then, we compare EVI-HRnet with existing facial video inpainting methods. 
Considering the lack of publicly available implementations of HMD removal methods that model temporal features in videos, we compare our results with existing facial video inpainting methods originally designed for moving mask removal across frames (our baseline model LGTSM ~\cite{chang2019learnable}, and CombCN ~\cite{wang2019video}). In CombCN, a two-stage deep video inpainting method, the temporal structure inference network is built on a 3D fully convolution architecture, and the spatial detail recovering network employs a 2D fully convolution network for image-based inpainting.

The section is structured as follows. We first describe the implementation details and setup of EVI-HRnet. Then, we introduce the dataset and the metrics which we use for the evaluation. Finally, we present and discuss the results of the ablation study and the comparison of EVI-HRnet with the baseline model LGTSM ~\cite{chang2019learnable} and CombCN ~\cite{wang2019video}.

\subsection{Implementation details and setup}

We utilize PyTorch 1.10.0 for the implementation of the network. We employ a kernel size of $5\times5$ for the first convolution layer, a kernel size of $4\times4$ with stride 2 for the down-sampling layers, a kernel size of $3\times3$ with dilation of 2, 4, 8, 16 for the dilated layers, and a kernel size of $3\times3$  for other convolution layers, similar to ~\cite{chang2019learnable}. For the attention layers, we utilize a kernel size of $1\times1$. As an activation function, we use the LeakyReLU. The Adam optimizer with a learning rate of $9.7 \times 10^{-5}$ is employed for training. 
Finally, we set the weights of the overall loss function (see Equation (\ref{equ:loss})) to $\lambda_{\text{AdvLoss}} = 1 $, $ \lambda_{\text{FERLoss}} = 2 $, $ \lambda_{\text{StyleLoss}} = 10 $, $ \lambda_{\text{VGGLoss}} = 1$, and $ \lambda_{\text{ReconLoss}} = 1$ for AdvLoss, FERLoss, StyleLoss, VGGLoss, and ReconLoss, respectively. 

\subsection{Video dataset}
To the best of our knowledge, there is a limited availability of facial video datasets compared to facial image datasets suitable for learning-based models. In this work, we employ the FaceForensics ~\cite{rossler2018faceforensics} dataset, which consists of 1,004 videos including more than 500,000 frames with faces of newscasters collected from YouTube. The majority of these videos contain frontal faces cropped to a size of $128\times128$ pixels, making it well-suited for training learning-based models. We use 150 of the videos for testing with a duration of 32 frames, while the remaining videos are allocated for training the models.

To simulate HMDs on the faces in the video frames, we first created binary masks of an HMD. Then, we applied the masks to all ground truth (GT) frames to be the input data for the inpainting networks. The HMD masks used for this study can be found in the supplementary material. 

\subsection{Evaluation metrics}
We use mean square error (MSE), peak-signal-to-noise ratio (PSNR),  and structural similarity index (SSIM) ~\cite{wang2004image} to evaluate the image quality in all models. Additionally, we also report Learned Perceptual Image Patch Similarity (LPIPS) ~\cite{zhang2018unreasonable}, and Fréchet inception distance (FID) ~\cite{heusel2017gans} score as our evaluation metrics, which have been demonstrated to align remarkably well with human judgments of image similarities.

MSE quantifies the overall pixel-level discrepancy between the two frames, with lower values indicating better similarity. PSNR provides a quantitative measure of how well the generated frame preserves the details and fidelity of the original frame. Higher PSNR values indicate better image quality. SSIM is a perception-based metric that evaluates the structural similarity between the generated frame and the ground truth. SSIM values range between -1 and 1, where 1 indicates perfect similarity. On the other hand, we use LPIPS, which is based on a learned feature representation of the images and provides a more human-centered evaluation of image quality. Lower LPIPS scores demonstrate higher perceptual similarity between the generated frame and the ground truth. Moreover, FID is utilized to evaluate the quality and temporal consistency of generated frames compared to real frames. Lower FID scores exhibit better video quality.

\subsection{Quantitative results}
\begin{figure}[t]
  \includegraphics[scale=0.4]{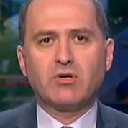}   
  \caption{Reference image (first frame) used for inpainting for a sequence (ID 78) in the FaceForensics testing set (image: MTV Lebanon News (\url{https://www.youtube.com/watch?v=nZJJVq_Mfvg})). 
  }
    \label{fig:ref_img}
\end{figure}

The quantitative evaluation results of EVI-HRnet, CombCN and LGTSM are presented in Table ~\ref{tab:metrics}. For EVI-HRNet, we are also interested to know the influence of landmarks detection and the FER loss in an ablation study. Thus, the table also shows the evaluation metrics of EVI-HRNet without the landmark detection and without the integration of the FER loss in Equation (\ref{equ:loss}).  
We can first observe that removing facial landmarks and the FER loss function has a negative effect on the model's overall performance. 
EVI-HRnet without facial landmarks or without the FER loss shows slightly higher MSE, LPIPS, and FID scores while having lower PSNR and SSIM scores compared to the proposed EVI-HRnet. These findings underscore the crucial roles played by both components in achieving realistic outputs with higher temporal consistency, indicated by a lower FID, and similarity. This indicates that using facial landmarks and the FER loss function improves the model's ability to generate realistic and temporally consistent frames. It is also noticeable that the FER loss has a higher contribution to the overall performance than the use of landmarks. In particular, the results of EVI-HRnet without FER loss have the highest FID, LPIPS, and MSE values, and lowest PSNR and SSIM values compared to the two other variations of EVI-HRnet, which indicates its considerable impact on generating realistic results. Removing the FER loss can thus lead to less realistic results.  

According to Table ~\ref{tab:metrics}, it is also apparent that the proposed EVI-HRnet performs better than CombCN and LGTSM based on the evaluation metrics values, achieving the lowest MSE, LPIPS, and FID scores, and the highest PSNR, and SSIM values. These results indicate a more satisfactory similarity between the inpainted and ground truth frames resulting from EVI-HRnet. With the lowest FID score, the proposed EVI-HRnet also exhibits higher temporal consistency in the generated videos.  

In summary, EVI-HRnet performs the best among all models and ablations in terms of the evaluation metrics values, indicating that it generates more realistic, similar, and consistent videos. The importance of facial landmarks and the FER loss is evident as removing them leads to a decrease in model performance. 


\begin{figure}[t]
    \centering
    \includegraphics[scale=0.3]{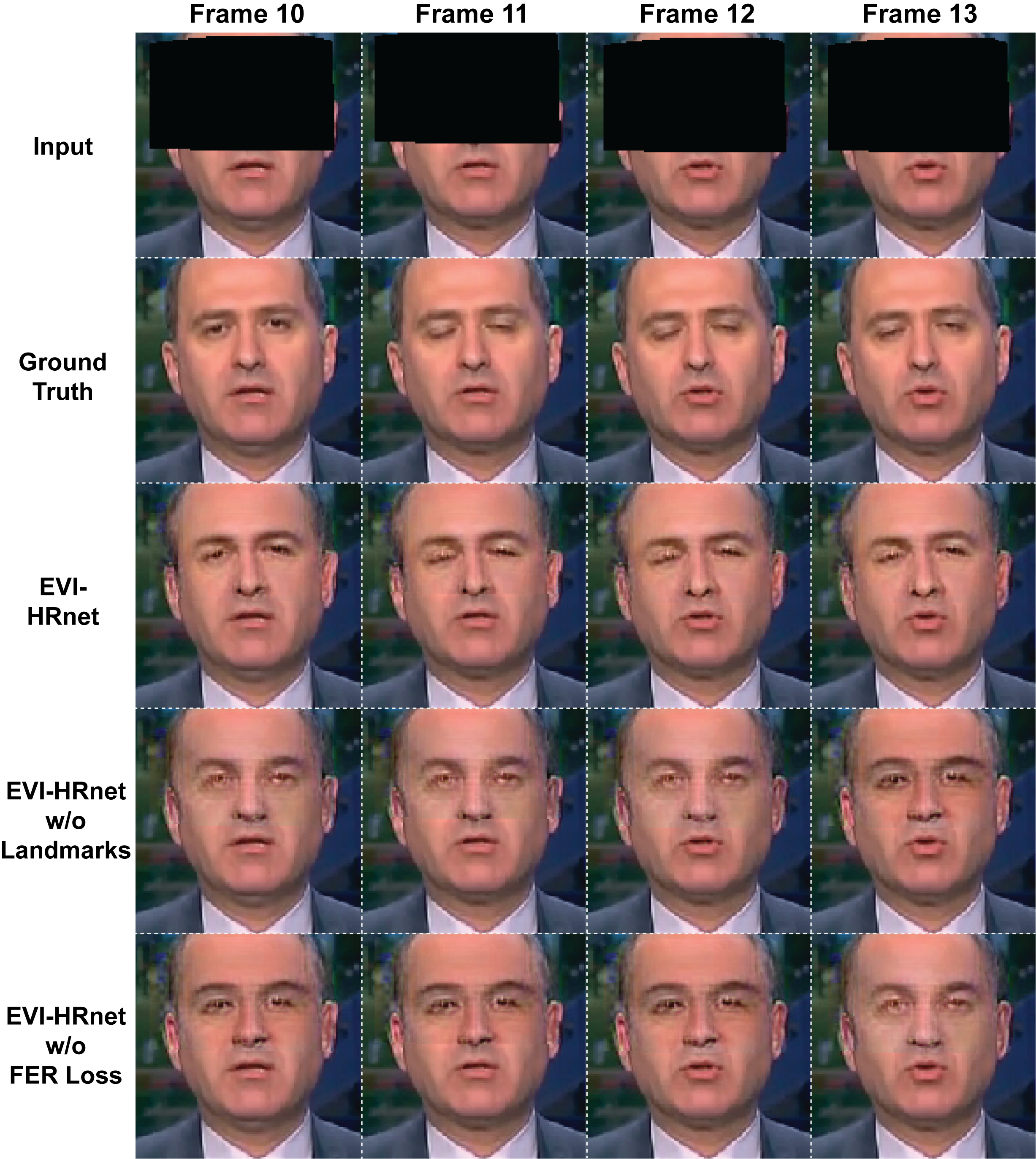}
    \caption{Sample of inpainted frames in FaceForensics validation set (ID 78) resulted from EVI-HRnet, EVI-HRnet without landmarks, and EVI-HRnet without the FER loss (images: MTV Lebanon News (\url{https://www.youtube.com/watch?v=nZJJVq_Mfvg})).}
    \label{fig:id78_abl}
\end{figure}


\begin{figure*}[t]
    \centering
    \includegraphics[width=0.7\linewidth]{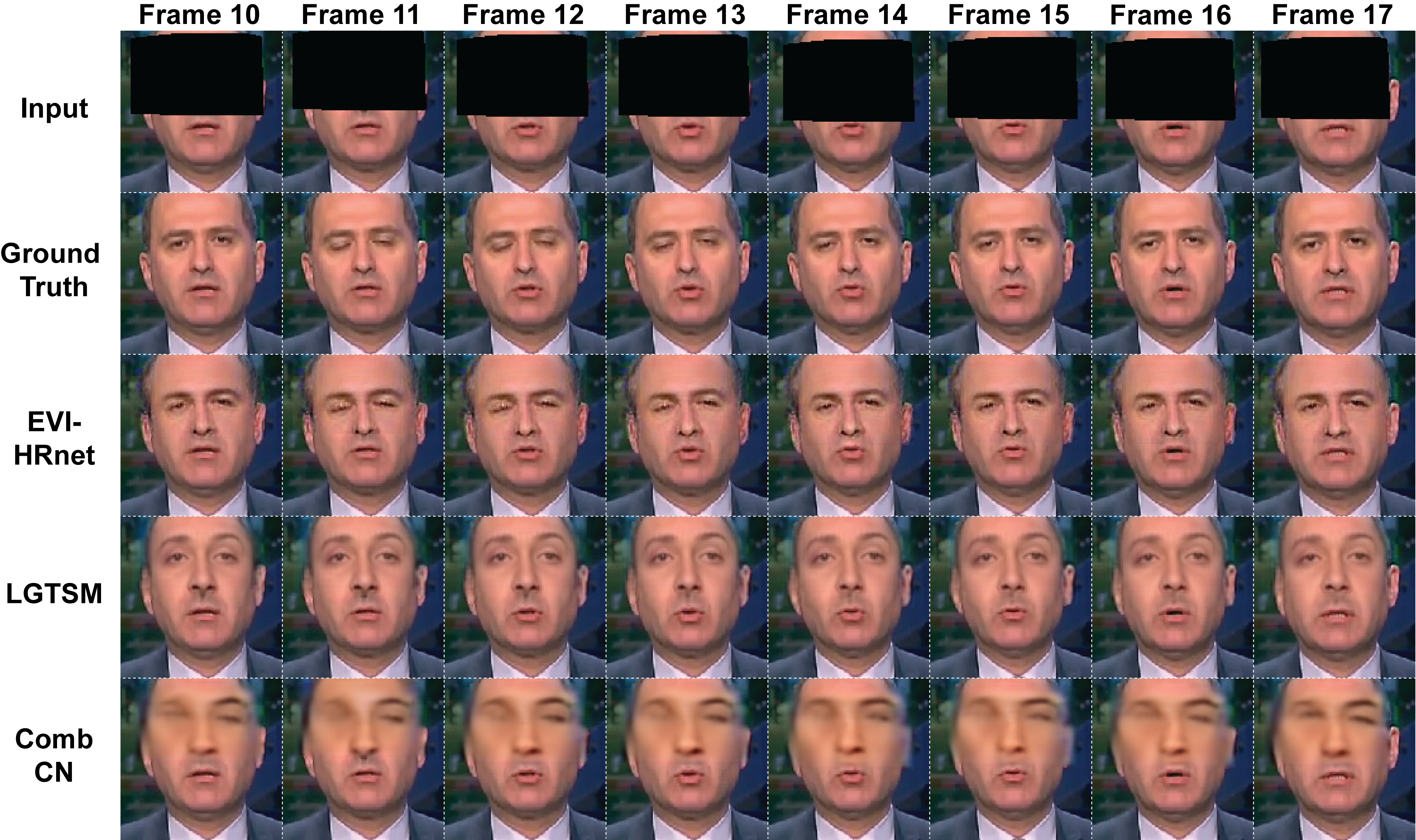}
    \caption{Sample of the qualitative results of FaceForensics validation set with HMD masks, and their GT and inputs (ID 78). The inpainted frames are selected from the results of EVI-HRnet, LGTSM, and CombCN (image: MTV Lebanon News (\url{https://www.youtube.com/watch?v=nZJJVq_Mfvg})).}
    \label{fig:quality_id78}
\end{figure*}

\subsection{Qualitative results}
Figure \ref{fig:id78_abl} shows exemplary inpainted frames using the proposed EVI-HRnet, EVI-HRnet without landmarks, and EVI-HRnet without FER loss together with the ground truth frames, and the masked inputs. The reference frame used by EVI-HRnet for the inpainting of this sequence is also illustrated in Figure ~\ref{fig:ref_img}. Note that only pixels of the reference images are used which are within the masked area of the input frames. 

As already indicated in the quantitative results, EVI-HRnet without facial landmarks performs better compared to EVI-HRnet without FER loss, which is especially noticeable at the eyelashes and eyebrows in Figure ~\ref{fig:id78_abl}. Moreover, it is apparent that EVI-HRnet complete model is more successful in preserving the natural flow of facial expressions (e.g. eye blinking) across frames and maintaining the authenticity of the subject's appearance. 

Samples of the qualitative results of EVI-HRnet, LGTSM, and CombCN are illustrated in Figure \ref{fig:quality_id78} for eight frames of the same test videos. Here, we exemplary selected the middle frames in each video as those are more challenging for the reconstruction networks compared to the first frames. The results are compared against the ground truth and the input frames for better comparison. 
Recovering fine details, such as subtle changes in facial expressions, eyebrow movements, and realistic eyelash rendering are vital in the HMD removal problem for creating a plausible experience in XR and visually appealing inpainted outputs. The results show that our proposed complete framework, EVI-HRnet, outperforms LGTSM and CombCN. With the help of a single reference image, FER loss, and facial landmarks, EVI-HRnet can generate visually realistic outputs by effectively addressing the HMD occlusion and preserving facial expressions. 
The recovered eye details and facial features, and smooth transition between expressions contribute to the overall realism and authenticity of the inpainted frames resulting from EVI-HRnet. This could enhance the viewer's perception and makes the inpainted frames more believable.

CombCN exhibited limitations in several instances, as it struggled to fully restore the eyes. In comparison, LGTSM displayed improved performance over CombCN by successfully recovering the eyes in every frame. However, a noticeable issue arises wherein the eyes are mispositioned, leading to inpainted frames that appear unchanged despite the varying facial expressions across frames present in the ground truth. Consequently, the failure to retain accurate identity and facial expressions stems from both networks' inability to access sufficient information from neighboring frames.  

Additional visual comparisons and videos with respect to the diversity of the subjects and reference frames used for inpainting can be found in the supplementary materials. 

\subsection{Discussion}
\label{sec:discussion}
The results demonstrate the good performance of EVI-HRnet in generating visually plausible results for HMD removal. Moreover, the results of the ablation study show the effectiveness of the FER loss in capturing essential expressions during the inpainting process.   

CombCN and LGTSM failed to recover the eyes and/or facial expressions in the occluded part of the faces. However, CombCN and LGTSM were originally developed for inpainting moving masks across frames where they achieve good performance as shown in \cite{wang2019video} and \cite{chang2019learnable}, respectively, since their primary focus is identifying missing features in neighboring frames during the learning process. By conducting this comparison, our aim was to highlight the existing research gap in current video inpainting methods, particularly when addressing the challenge of HMD removal where the mask is not moving.

It is further important to acknowledge the potential limitations of our approach to providing a comprehensive evaluation of EVI-HRnet. One notable limitation is related to the training data's distribution and the lack of diverse facial videos with extreme head movements or extreme expression changes. The majority of the training data comprises frontal facial views, which can result in minor imperfections when reconstructing complex facial features, such as the eyes, especially in situations involving significant head movements or rapid expression shifts.

To mitigate this limitation and further enhance the model's performance, a more extensive range of facial views, expressions, and head movements in the dataset is required. Incorporating such diverse data would expose the model to a broader set of challenges, enabling it to learn more robust and accurate representations of various facial features, including the eyes.

\begin{figure}[t] 
  \subfigure{%
    \includegraphics[scale=0.3]{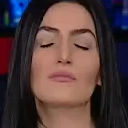}  
  } 
  \subfigure{%
    \includegraphics[scale=0.3]{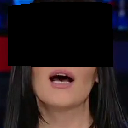}   
  }
  \subfigure{%
    \includegraphics[scale=0.3]{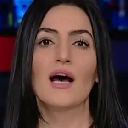}   
  }
  \subfigure{%
    \includegraphics[scale=0.3]{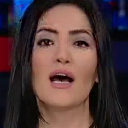}   
  }

  \caption{Examples of inpainted results of reference images with closed eyes for a sequence (ID 10) in the FaceForensics testing set. From left to right: reference frame, input, GT, and result from EVI-HRnet (image: Al Aan TV (\url{https://www.youtube.com/watch?v=2iBnQpy8OMw})).
  }
    \label{fig:no-eye}
\end{figure}

Furthermore, while EVI-HRnet demonstrates proficiency in restoring facial features in frontal views, even when the eyes are closed in the reference frame (see Figure~\ref{fig:no-eye}), the task becomes more challenging when dealing with non-frontal perspectives. The reason for the latter is that the accuracy of detected facial landmarks and expressions diminishes in such scenarios. Additionally, the lack of a high-quality reference image featuring open eyes adds to the complexity of the inpainting process. Despite the absence of eye features in the reference image (ID 10 with a frontal view), EVI-HRnet still could produce relatively realistic outputs, which shows the efficiency of the FER loss and the facial landmarks detected from the original frames.

On the other hand, it is worth noting that facial landmarks themselves may have inherent limitations. The accuracy of facial landmark detection heavily depends on the quality of the input image and the complexity of facial expressions. In our experiments, we applied landmark detection on the ground truth data. In a real scenario, landmarks, which are covered by the HMD, need to be captured with internal cameras of the HMD, which likely degrades the accuracy of the detection algorithm. Hence, in scenarios where facial landmarks are not accurately detected, the inpainting process could be adversely affected, leading to less accurate results.
\section{Conclusion and future work}
\label{sec:conclusion}
HMDs present a major barrier to a pleasant user experience by blocking the upper face of the user in social and collaborative XR applications. In this study, we proposed EVI-HRnet for HMD removal in videos. EVI-HRnet effectively inpainted missing facial information by leveraging facial landmarks and a single occlusion-free reference image of the user. We have also introduced a novel FER loss function for enhancing the quality and realism of the inpainted faces. Through both qualitative and quantitative experiments, we validated the effectiveness of each component and demonstrated the framework's capability to accurately recover real emotions during HMD removal. 

Despite the promising results, several challenges still remain, e.g. to accurately recover occluded parts and maintain temporal consistency in scenarios involving significant head movements or sudden expression changes. Thus, investigations and refinements are necessary to fully unleash the framework's capabilities and realize its potential and challenges in real-world settings. Beyond the current proof-of-concept of the proposed pipeline, testing the approach with HMDs equipped with internal cameras for facial landmark detection would be an essential step toward validating its real-world applicability and robustness. 

Furthermore, it is worth mentioning the potential application of 3D reconstruction from a single view to obtain 3D models of faces for XR applications. Although this aspect has not been addressed in the current work, it represents a vital element in social XR applications, as 2D videos may not fully exploit the immersive potential of XR environments. Exploring the integration of 3D facial models could open up new possibilities for more immersive and realistic social interactions within XR environments.
 
 Future research could also explore the incorporation of multimodal video inpainting techniques, which may improve robustness for handling large head movements and complex facial dynamics.

\begin{acks}
This project has received funding from the European Union’s Horizon 2020 research and innovation program under the Marie Skłodowska-Curie grant agreement No 956770. 
\end{acks}
\bibliographystyle{ACM-Reference-Format}
\bibliography{refs}


\begin{thebibliography}{44}


\ifx \showCODEN    \undefined \def \showCODEN     #1{\unskip}     \fi
\ifx \showDOI      \undefined \def \showDOI       #1{#1}\fi
\ifx \showISBNx    \undefined \def \showISBNx     #1{\unskip}     \fi
\ifx \showISBNxiii \undefined \def \showISBNxiii  #1{\unskip}     \fi
\ifx \showISSN     \undefined \def \showISSN      #1{\unskip}     \fi
\ifx \showLCCN     \undefined \def \showLCCN      #1{\unskip}     \fi
\ifx \shownote     \undefined \def \shownote      #1{#1}          \fi
\ifx \showarticletitle \undefined \def \showarticletitle #1{#1}   \fi
\ifx \showURL      \undefined \def \showURL       {\relax}        \fi
\providecommand\bibfield[2]{#2}
\providecommand\bibinfo[2]{#2}
\providecommand\natexlab[1]{#1}
\providecommand\showeprint[2][]{arXiv:#2}

\bibitem[Ali et~al\mbox{.}(2023)]%
        {ali2023systematic}
\bibfield{author}{\bibinfo{person}{Saba~Ghazanfar Ali},
  \bibinfo{person}{Xiangning Wang}, \bibinfo{person}{Ping Li},
  \bibinfo{person}{Younhyun Jung}, \bibinfo{person}{Lei Bi},
  \bibinfo{person}{Jinman Kim}, \bibinfo{person}{Yuting Chen},
  \bibinfo{person}{David~Dagan Feng}, \bibinfo{person}{Nadia
  Magnenat~Thalmann}, \bibinfo{person}{Jihong Wang}, {et~al\mbox{.}}}
  \bibinfo{year}{2023}\natexlab{}.
\newblock \showarticletitle{A systematic review: Virtual-reality-based
  techniques for human exercises and health improvement}.
\newblock \bibinfo{journal}{\emph{Frontiers in Public Health}}
  \bibinfo{volume}{11} (\bibinfo{year}{2023}), \bibinfo{pages}{1143947}.
\newblock


\bibitem[Arjovsky et~al\mbox{.}(2017)]%
        {arjovsky2017wasserstein}
\bibfield{author}{\bibinfo{person}{Martin Arjovsky}, \bibinfo{person}{Soumith
  Chintala}, {and} \bibinfo{person}{L{\'e}on Bottou}.}
  \bibinfo{year}{2017}\natexlab{}.
\newblock \showarticletitle{Wasserstein generative adversarial networks}. In
  \bibinfo{booktitle}{\emph{International conference on machine learning}}.
  PMLR, \bibinfo{pages}{214--223}.
\newblock


\bibitem[Chang et~al\mbox{.}(2019a)]%
        {chang2019free}
\bibfield{author}{\bibinfo{person}{Ya-Liang Chang}, \bibinfo{person}{Zhe~Yu
  Liu}, \bibinfo{person}{Kuan-Ying Lee}, {and} \bibinfo{person}{Winston Hsu}.}
  \bibinfo{year}{2019}\natexlab{a}.
\newblock \showarticletitle{Free-form video inpainting with 3d gated
  convolution and temporal patchgan}. In \bibinfo{booktitle}{\emph{Proceedings
  of the IEEE/CVF International Conference on Computer Vision}}.
  \bibinfo{pages}{9066--9075}.
\newblock


\bibitem[Chang et~al\mbox{.}(2019b)]%
        {chang2019learnable}
\bibfield{author}{\bibinfo{person}{Ya-Liang Chang}, \bibinfo{person}{Zhe~Yu
  Liu}, \bibinfo{person}{Kuan-Ying Lee}, {and} \bibinfo{person}{Winston Hsu}.}
  \bibinfo{year}{2019}\natexlab{b}.
\newblock \showarticletitle{Learnable gated temporal shift module for deep
  video inpainting}.
\newblock \bibinfo{journal}{\emph{arXiv preprint arXiv:1907.01131}}
  (\bibinfo{year}{2019}).
\newblock


\bibitem[Chen et~al\mbox{.}(2023b)]%
        {chen2023multi}
\bibfield{author}{\bibinfo{person}{Gang Chen}, \bibinfo{person}{Guipeng Zhang},
  \bibinfo{person}{Zhenguo Yang}, {and} \bibinfo{person}{Wenyin Liu}.}
  \bibinfo{year}{2023}\natexlab{b}.
\newblock \showarticletitle{Multi-scale patch-GAN with edge detection for image
  inpainting}.
\newblock \bibinfo{journal}{\emph{Applied Intelligence}} \bibinfo{volume}{53},
  \bibinfo{number}{4} (\bibinfo{year}{2023}), \bibinfo{pages}{3917--3932}.
\newblock


\bibitem[Chen et~al\mbox{.}(2018)]%
        {chen2018real}
\bibfield{author}{\bibinfo{person}{Shu-Yu Chen}, \bibinfo{person}{Lin Gao},
  \bibinfo{person}{Yu-Kun Lai}, \bibinfo{person}{Paul~L Rosin}, {and}
  \bibinfo{person}{Shihong Xia}.} \bibinfo{year}{2018}\natexlab{}.
\newblock \showarticletitle{Real-time 3d face reconstruction and gaze tracking
  for virtual reality}. In \bibinfo{booktitle}{\emph{2018 IEEE Conference on
  virtual reality and 3d user interfaces (VR)}}. IEEE,
  \bibinfo{pages}{525--526}.
\newblock


\bibitem[Chen et~al\mbox{.}(2022)]%
        {chen20223d}
\bibfield{author}{\bibinfo{person}{Shu-Yu Chen}, \bibinfo{person}{Yu-Kun Lai},
  \bibinfo{person}{Shihong Xia}, \bibinfo{person}{Paul Rosin}, {and}
  \bibinfo{person}{Lin Gao}.} \bibinfo{year}{2022}\natexlab{}.
\newblock \showarticletitle{3D face reconstruction and gaze tracking in the HMD
  for virtual interaction}.
\newblock \bibinfo{journal}{\emph{IEEE Transactions on Multimedia}}
  (\bibinfo{year}{2022}).
\newblock


\bibitem[Chen et~al\mbox{.}(2023a)]%
        {chen2023dgca}
\bibfield{author}{\bibinfo{person}{Yuantao Chen}, \bibinfo{person}{Runlong
  Xia}, \bibinfo{person}{Kai Yang}, {and} \bibinfo{person}{Ke Zou}.}
  \bibinfo{year}{2023}\natexlab{a}.
\newblock \showarticletitle{DGCA: high resolution image inpainting via DR-GAN
  and contextual attention}.
\newblock \bibinfo{journal}{\emph{Multimedia Tools and Applications}}
  (\bibinfo{year}{2023}), \bibinfo{pages}{1--21}.
\newblock


\bibitem[Dhall et~al\mbox{.}(2012)]%
        {dhall2012collecting}
\bibfield{author}{\bibinfo{person}{Abhinav Dhall}, \bibinfo{person}{Roland
  Goecke}, \bibinfo{person}{Simon Lucey}, \bibinfo{person}{Tom Gedeon},
  {et~al\mbox{.}}} \bibinfo{year}{2012}\natexlab{}.
\newblock \showarticletitle{Collecting large, richly annotated
  facial-expression databases from movies}.
\newblock \bibinfo{journal}{\emph{IEEE multimedia}} \bibinfo{volume}{19},
  \bibinfo{number}{3} (\bibinfo{year}{2012}), \bibinfo{pages}{34}.
\newblock


\bibitem[Frueh et~al\mbox{.}(2017)]%
        {10.1145/3084363.3085083}
\bibfield{author}{\bibinfo{person}{Christian Frueh}, \bibinfo{person}{Avneesh
  Sud}, {and} \bibinfo{person}{Vivek Kwatra}.} \bibinfo{year}{2017}\natexlab{}.
\newblock \showarticletitle{Headset Removal for Virtual and Mixed Reality}. In
  \bibinfo{booktitle}{\emph{ACM SIGGRAPH 2017 Talks}} (Los Angeles, California)
  \emph{(\bibinfo{series}{SIGGRAPH '17})}. \bibinfo{publisher}{Association for
  Computing Machinery}, \bibinfo{address}{New York, NY, USA}, Article
  \bibinfo{articleno}{80}, \bibinfo{numpages}{2}~pages.
\newblock
\showISBNx{9781450350082}
\urldef\tempurl%
\url{https://doi.org/10.1145/3084363.3085083}
\showDOI{\tempurl}


\bibitem[Gatys et~al\mbox{.}(2015)]%
        {gatys2015neural}
\bibfield{author}{\bibinfo{person}{Leon~A Gatys}, \bibinfo{person}{Alexander~S
  Ecker}, {and} \bibinfo{person}{Matthias Bethge}.}
  \bibinfo{year}{2015}\natexlab{}.
\newblock \showarticletitle{A neural algorithm of artistic style}.
\newblock \bibinfo{journal}{\emph{arXiv preprint arXiv:1508.06576}}
  (\bibinfo{year}{2015}).
\newblock


\bibitem[Goodfellow et~al\mbox{.}(2014)]%
        {goodfellow2014generative}
\bibfield{author}{\bibinfo{person}{Ian Goodfellow}, \bibinfo{person}{Jean
  Pouget-Abadie}, \bibinfo{person}{Mehdi Mirza}, \bibinfo{person}{Bing Xu},
  \bibinfo{person}{David Warde-Farley}, \bibinfo{person}{Sherjil Ozair},
  \bibinfo{person}{Aaron Courville}, {and} \bibinfo{person}{Yoshua Bengio}.}
  \bibinfo{year}{2014}\natexlab{}.
\newblock \showarticletitle{Generative adversarial nets}.
\newblock \bibinfo{journal}{\emph{Advances in neural information processing
  systems}}  \bibinfo{volume}{27} (\bibinfo{year}{2014}).
\newblock


\bibitem[Gupta et~al\mbox{.}(2022a)]%
        {gupta2022supervision}
\bibfield{author}{\bibinfo{person}{Surabhi Gupta}, \bibinfo{person}{Sai~Sagar
  Jinka}, \bibinfo{person}{Avinash Sharma}, {and} \bibinfo{person}{Anoop
  Namboodiri}.} \bibinfo{year}{2022}\natexlab{a}.
\newblock \showarticletitle{Supervision by Landmarks: An Enhanced Facial
  De-occlusion Network for VR-Based Applications}. In
  \bibinfo{booktitle}{\emph{European Conference on Computer Vision}}. Springer,
  \bibinfo{pages}{323--337}.
\newblock


\bibitem[Gupta et~al\mbox{.}(2022b)]%
        {gupta2022attention}
\bibfield{author}{\bibinfo{person}{Surabhi Gupta}, \bibinfo{person}{Ashwath
  Shetty}, {and} \bibinfo{person}{Avinash Sharma}.}
  \bibinfo{year}{2022}\natexlab{b}.
\newblock \showarticletitle{Attention based occlusion removal for hybrid
  telepresence systems}. In \bibinfo{booktitle}{\emph{2022 19th Conference on
  Robots and Vision (CRV)}}. IEEE, \bibinfo{pages}{167--174}.
\newblock


\bibitem[Heusel et~al\mbox{.}(2017)]%
        {heusel2017gans}
\bibfield{author}{\bibinfo{person}{Martin Heusel}, \bibinfo{person}{Hubert
  Ramsauer}, \bibinfo{person}{Thomas Unterthiner}, \bibinfo{person}{Bernhard
  Nessler}, {and} \bibinfo{person}{Sepp Hochreiter}.}
  \bibinfo{year}{2017}\natexlab{}.
\newblock \showarticletitle{Gans trained by a two time-scale update rule
  converge to a local nash equilibrium}.
\newblock \bibinfo{journal}{\emph{Advances in neural information processing
  systems}}  \bibinfo{volume}{30} (\bibinfo{year}{2017}).
\newblock


\bibitem[Kazemi and Sullivan(2014)]%
        {kazemi2014one}
\bibfield{author}{\bibinfo{person}{Vahid Kazemi} {and}
  \bibinfo{person}{Josephine Sullivan}.} \bibinfo{year}{2014}\natexlab{}.
\newblock \showarticletitle{One millisecond face alignment with an ensemble of
  regression trees}. In \bibinfo{booktitle}{\emph{Proceedings of the IEEE
  conference on computer vision and pattern recognition}}.
  \bibinfo{pages}{1867--1874}.
\newblock


\bibitem[King(2009)]%
        {dlib09}
\bibfield{author}{\bibinfo{person}{Davis~E. King}.}
  \bibinfo{year}{2009}\natexlab{}.
\newblock \showarticletitle{Dlib-ml: A Machine Learning Toolkit}.
\newblock \bibinfo{journal}{\emph{Journal of Machine Learning Research}}
  \bibinfo{volume}{10} (\bibinfo{year}{2009}), \bibinfo{pages}{1755--1758}.
\newblock


\bibitem[Lahiri et~al\mbox{.}(2020)]%
        {lahiri2020prior}
\bibfield{author}{\bibinfo{person}{Avisek Lahiri}, \bibinfo{person}{Arnav~Kumar
  Jain}, \bibinfo{person}{Sanskar Agrawal}, \bibinfo{person}{Pabitra Mitra},
  {and} \bibinfo{person}{Prabir~Kumar Biswas}.}
  \bibinfo{year}{2020}\natexlab{}.
\newblock \showarticletitle{Prior guided gan based semantic inpainting}. In
  \bibinfo{booktitle}{\emph{Proceedings of the IEEE/CVF conference on computer
  vision and pattern recognition}}. \bibinfo{pages}{13696--13705}.
\newblock


\bibitem[Langa et~al\mbox{.}(2022)]%
        {9848469}
\bibfield{author}{\bibinfo{person}{Sergi~Fernández Langa},
  \bibinfo{person}{Mario Montagud}, \bibinfo{person}{Gianluca Cernigliaro},
  {and} \bibinfo{person}{David~Rincón Rivera}.}
  \bibinfo{year}{2022}\natexlab{}.
\newblock \showarticletitle{Multiparty Holomeetings: Toward a New Era of
  Low-Cost Volumetric Holographic Meetings in Virtual Reality}.
\newblock \bibinfo{journal}{\emph{IEEE Access}}  \bibinfo{volume}{10}
  (\bibinfo{year}{2022}), \bibinfo{pages}{81856--81876}.
\newblock
\urldef\tempurl%
\url{https://doi.org/10.1109/ACCESS.2022.3196285}
\showDOI{\tempurl}


\bibitem[Lin et~al\mbox{.}(2019)]%
        {lin2019tsm}
\bibfield{author}{\bibinfo{person}{Ji Lin}, \bibinfo{person}{Chuang Gan}, {and}
  \bibinfo{person}{Song Han}.} \bibinfo{year}{2019}\natexlab{}.
\newblock \showarticletitle{Tsm: Temporal shift module for efficient video
  understanding}. In \bibinfo{booktitle}{\emph{Proceedings of the IEEE/CVF
  international conference on computer vision}}. \bibinfo{pages}{7083--7093}.
\newblock


\bibitem[Lou et~al\mbox{.}(2019)]%
        {lou2019realistic}
\bibfield{author}{\bibinfo{person}{Jianwen Lou}, \bibinfo{person}{Yiming Wang},
  \bibinfo{person}{Charles Nduka}, \bibinfo{person}{Mahyar Hamedi},
  \bibinfo{person}{Ifigeneia Mavridou}, \bibinfo{person}{Fei-Yue Wang}, {and}
  \bibinfo{person}{Hui Yu}.} \bibinfo{year}{2019}\natexlab{}.
\newblock \showarticletitle{Realistic facial expression reconstruction for VR
  HMD users}.
\newblock \bibinfo{journal}{\emph{IEEE Transactions on Multimedia}}
  \bibinfo{volume}{22}, \bibinfo{number}{3} (\bibinfo{year}{2019}),
  \bibinfo{pages}{730--743}.
\newblock


\bibitem[Numan(2020)]%
        {numan2020generative}
\bibfield{author}{\bibinfo{person}{Nels Numan}.}
  \bibinfo{year}{2020}\natexlab{}.
\newblock \bibinfo{title}{Generative RGB-D Face Completion for Head-Mounted
  Display Removal}.
\newblock
\newblock
\urldef\tempurl%
\url{http://resolver.tudelft.nl/uuid:ed0680e5-ea0c-49e3-a1a1-be9de8773188}
\showURL{%
\tempurl}


\bibitem[Numan et~al\mbox{.}(2021)]%
        {numan2021generative}
\bibfield{author}{\bibinfo{person}{Nels Numan}, \bibinfo{person}{Frank
  Ter~Haar}, {and} \bibinfo{person}{Pablo Cesar}.}
  \bibinfo{year}{2021}\natexlab{}.
\newblock \showarticletitle{Generative RGB-D face completion for head-mounted
  display removal}. In \bibinfo{booktitle}{\emph{2021 IEEE Conference on
  Virtual Reality and 3D User Interfaces Abstracts and Workshops (VRW)}}. IEEE,
  \bibinfo{pages}{109--116}.
\newblock


\bibitem[Prins et~al\mbox{.}(2018)]%
        {prins2018togethervr}
\bibfield{author}{\bibinfo{person}{Martin~J Prins}, \bibinfo{person}{Simon~NB
  Gunkel}, \bibinfo{person}{Hans~M Stokking}, {and} \bibinfo{person}{Omar~A
  Niamut}.} \bibinfo{year}{2018}\natexlab{}.
\newblock \showarticletitle{TogetherVR: A framework for photorealistic shared
  media experiences in 360-degree VR}.
\newblock \bibinfo{journal}{\emph{SMPTE Motion Imaging Journal}}
  \bibinfo{volume}{127}, \bibinfo{number}{7} (\bibinfo{year}{2018}),
  \bibinfo{pages}{39--44}.
\newblock


\bibitem[Rojas-S{\'a}nchez et~al\mbox{.}(2023)]%
        {rojas2023systematic}
\bibfield{author}{\bibinfo{person}{Mario~A Rojas-S{\'a}nchez},
  \bibinfo{person}{Pedro~R Palos-S{\'a}nchez}, {and}
  \bibinfo{person}{Jos{\'e}~A Folgado-Fern{\'a}ndez}.}
  \bibinfo{year}{2023}\natexlab{}.
\newblock \showarticletitle{Systematic literature review and bibliometric
  analysis on virtual reality and education}.
\newblock \bibinfo{journal}{\emph{Education and Information Technologies}}
  \bibinfo{volume}{28}, \bibinfo{number}{1} (\bibinfo{year}{2023}),
  \bibinfo{pages}{155--192}.
\newblock


\bibitem[R{\"o}ssler et~al\mbox{.}(2018)]%
        {rossler2018faceforensics}
\bibfield{author}{\bibinfo{person}{Andreas R{\"o}ssler},
  \bibinfo{person}{Davide Cozzolino}, \bibinfo{person}{Luisa Verdoliva},
  \bibinfo{person}{Christian Riess}, \bibinfo{person}{Justus Thies}, {and}
  \bibinfo{person}{Matthias Nie{\ss}ner}.} \bibinfo{year}{2018}\natexlab{}.
\newblock \showarticletitle{Faceforensics: A large-scale video dataset for
  forgery detection in human faces}.
\newblock \bibinfo{journal}{\emph{arXiv preprint arXiv:1803.09179}}
  (\bibinfo{year}{2018}).
\newblock


\bibitem[Russakovsky et~al\mbox{.}(2015)]%
        {russakovsky2015imagenet}
\bibfield{author}{\bibinfo{person}{Olga Russakovsky}, \bibinfo{person}{Jia
  Deng}, \bibinfo{person}{Hao Su}, \bibinfo{person}{Jonathan Krause},
  \bibinfo{person}{Sanjeev Satheesh}, \bibinfo{person}{Sean Ma},
  \bibinfo{person}{Zhiheng Huang}, \bibinfo{person}{Andrej Karpathy},
  \bibinfo{person}{Aditya Khosla}, \bibinfo{person}{Michael Bernstein},
  {et~al\mbox{.}}} \bibinfo{year}{2015}\natexlab{}.
\newblock \showarticletitle{Imagenet large scale visual recognition challenge}.
\newblock \bibinfo{journal}{\emph{International journal of computer vision}}
  \bibinfo{volume}{115} (\bibinfo{year}{2015}), \bibinfo{pages}{211--252}.
\newblock


\bibitem[Savchenko(2022)]%
        {savchenko2022video}
\bibfield{author}{\bibinfo{person}{Andrey~V Savchenko}.}
  \bibinfo{year}{2022}\natexlab{}.
\newblock \showarticletitle{Video-based frame-level facial analysis of
  affective behavior on mobile devices using EfficientNets}. In
  \bibinfo{booktitle}{\emph{Proceedings of the IEEE/CVF Conference on Computer
  Vision and Pattern Recognition}}. \bibinfo{pages}{2359--2366}.
\newblock


\bibitem[Simons et~al\mbox{.}(2023)]%
        {simons2023intelligence}
\bibfield{author}{\bibinfo{person}{Alexander Simons}, \bibinfo{person}{Isabell
  Wohlgenannt}, \bibinfo{person}{Sarah Zelt}, \bibinfo{person}{Markus
  Weinmann}, \bibinfo{person}{Johannes Schneider}, {and} \bibinfo{person}{Jan
  vom Brocke}.} \bibinfo{year}{2023}\natexlab{}.
\newblock \showarticletitle{Intelligence at play: game-based assessment using a
  virtual-reality application}.
\newblock \bibinfo{journal}{\emph{Virtual Reality}} (\bibinfo{year}{2023}),
  \bibinfo{pages}{1--17}.
\newblock


\bibitem[Simonyan and Zisserman(2014)]%
        {simonyan2014very}
\bibfield{author}{\bibinfo{person}{Karen Simonyan} {and}
  \bibinfo{person}{Andrew Zisserman}.} \bibinfo{year}{2014}\natexlab{}.
\newblock \showarticletitle{Very deep convolutional networks for large-scale
  image recognition}.
\newblock \bibinfo{journal}{\emph{arXiv preprint arXiv:1409.1556}}
  (\bibinfo{year}{2014}).
\newblock


\bibitem[Szeto and Corso(2022)]%
        {szeto2022devil}
\bibfield{author}{\bibinfo{person}{Ryan Szeto} {and} \bibinfo{person}{Jason~J
  Corso}.} \bibinfo{year}{2022}\natexlab{}.
\newblock \showarticletitle{The devil is in the details: A diagnostic
  evaluation benchmark for video inpainting}. In
  \bibinfo{booktitle}{\emph{Proceedings of the IEEE/CVF Conference on Computer
  Vision and Pattern Recognition}}. \bibinfo{pages}{21054--21063}.
\newblock


\bibitem[Thies et~al\mbox{.}(2016)]%
        {thies2016facevr}
\bibfield{author}{\bibinfo{person}{Justus Thies}, \bibinfo{person}{Michael
  Zollh{\"o}fer}, \bibinfo{person}{Marc Stamminger}, \bibinfo{person}{Christian
  Theobalt}, {and} \bibinfo{person}{Matthias Nie{\ss}ner}.}
  \bibinfo{year}{2016}\natexlab{}.
\newblock \showarticletitle{Facevr: Real-time facial reenactment and eye gaze
  control in virtual reality}.
\newblock \bibinfo{journal}{\emph{arXiv preprint arXiv:1610.03151}}
  (\bibinfo{year}{2016}).
\newblock


\bibitem[Van Der~Boon et~al\mbox{.}(2022)]%
        {van2022deep}
\bibfield{author}{\bibinfo{person}{Matthijs Van Der~Boon},
  \bibinfo{person}{Leonor Fermoselle}, \bibinfo{person}{Frank Ter~Haar},
  \bibinfo{person}{Sylvie Dijkstra-Soudarissanane}, {and} \bibinfo{person}{Omar
  Niamut}.} \bibinfo{year}{2022}\natexlab{}.
\newblock \showarticletitle{Deep Learning Augmented Realistic Avatars for
  Social VR Human Representation}. In \bibinfo{booktitle}{\emph{ACM
  International Conference on Interactive Media Experiences}}.
  \bibinfo{pages}{311--318}.
\newblock


\bibitem[Wang et~al\mbox{.}(2019a)]%
        {wang2019video}
\bibfield{author}{\bibinfo{person}{Chuan Wang}, \bibinfo{person}{Haibin Huang},
  \bibinfo{person}{Xiaoguang Han}, {and} \bibinfo{person}{Jue Wang}.}
  \bibinfo{year}{2019}\natexlab{a}.
\newblock \showarticletitle{Video inpainting by jointly learning temporal
  structure and spatial details}. In \bibinfo{booktitle}{\emph{Proceedings of
  the AAAI Conference on Artificial Intelligence}}, Vol.~\bibinfo{volume}{33}.
  \bibinfo{pages}{5232--5239}.
\newblock


\bibitem[Wang et~al\mbox{.}(2019b)]%
        {wang2019faithful}
\bibfield{author}{\bibinfo{person}{Miao Wang}, \bibinfo{person}{Xin Wen}, {and}
  \bibinfo{person}{Shi-Min Hu}.} \bibinfo{year}{2019}\natexlab{b}.
\newblock \showarticletitle{Faithful face image completion for HMD occlusion
  removal}. In \bibinfo{booktitle}{\emph{2019 IEEE International Symposium on
  Mixed and Augmented Reality Adjunct (ISMAR-Adjunct)}}. IEEE,
  \bibinfo{pages}{251--256}.
\newblock


\bibitem[Wang et~al\mbox{.}(2004)]%
        {wang2004image}
\bibfield{author}{\bibinfo{person}{Zhou Wang}, \bibinfo{person}{Alan~C Bovik},
  \bibinfo{person}{Hamid~R Sheikh}, {and} \bibinfo{person}{Eero~P Simoncelli}.}
  \bibinfo{year}{2004}\natexlab{}.
\newblock \showarticletitle{Image quality assessment: from error visibility to
  structural similarity}.
\newblock \bibinfo{journal}{\emph{IEEE transactions on image processing}}
  \bibinfo{volume}{13}, \bibinfo{number}{4} (\bibinfo{year}{2004}),
  \bibinfo{pages}{600--612}.
\newblock


\bibitem[Wu et~al\mbox{.}(2020)]%
        {wu2020image}
\bibfield{author}{\bibinfo{person}{Yifan Wu}, \bibinfo{person}{Vivek Singh},
  {and} \bibinfo{person}{Ankur Kapoor}.} \bibinfo{year}{2020}\natexlab{}.
\newblock \showarticletitle{From image to video face inpainting:
  spatial-temporal nested GAN (STN-GAN) for usability recovery}. In
  \bibinfo{booktitle}{\emph{Proceedings of the IEEE/CVF Winter Conference on
  Applications of Computer Vision}}. \bibinfo{pages}{2396--2405}.
\newblock


\bibitem[Yang et~al\mbox{.}(2023)]%
        {yang2023deep}
\bibfield{author}{\bibinfo{person}{Wenqi Yang}, \bibinfo{person}{Zhenfang
  Chen}, \bibinfo{person}{Chaofeng Chen}, \bibinfo{person}{Guanying Chen},
  {and} \bibinfo{person}{Kwan-Yee~K Wong}.} \bibinfo{year}{2023}\natexlab{}.
\newblock \showarticletitle{Deep Face Video Inpainting via UV Mapping}.
\newblock \bibinfo{journal}{\emph{IEEE Transactions on Image Processing}}
  \bibinfo{volume}{32} (\bibinfo{year}{2023}), \bibinfo{pages}{1145--1157}.
\newblock


\bibitem[Yu et~al\mbox{.}(2019)]%
        {yu2019free}
\bibfield{author}{\bibinfo{person}{Jiahui Yu}, \bibinfo{person}{Zhe Lin},
  \bibinfo{person}{Jimei Yang}, \bibinfo{person}{Xiaohui Shen},
  \bibinfo{person}{Xin Lu}, {and} \bibinfo{person}{Thomas~S Huang}.}
  \bibinfo{year}{2019}\natexlab{}.
\newblock \showarticletitle{Free-form image inpainting with gated convolution}.
  In \bibinfo{booktitle}{\emph{Proceedings of the IEEE/CVF international
  conference on computer vision}}. \bibinfo{pages}{4471--4480}.
\newblock


\bibitem[Zhang et~al\mbox{.}(2019)]%
        {zhang2019self}
\bibfield{author}{\bibinfo{person}{Han Zhang}, \bibinfo{person}{Ian
  Goodfellow}, \bibinfo{person}{Dimitris Metaxas}, {and}
  \bibinfo{person}{Augustus Odena}.} \bibinfo{year}{2019}\natexlab{}.
\newblock \showarticletitle{Self-attention generative adversarial networks}. In
  \bibinfo{booktitle}{\emph{International conference on machine learning}}.
  PMLR, \bibinfo{pages}{7354--7363}.
\newblock


\bibitem[Zhang et~al\mbox{.}(2016)]%
        {zhang2016joint}
\bibfield{author}{\bibinfo{person}{Kaipeng Zhang}, \bibinfo{person}{Zhanpeng
  Zhang}, \bibinfo{person}{Zhifeng Li}, {and} \bibinfo{person}{Yu Qiao}.}
  \bibinfo{year}{2016}\natexlab{}.
\newblock \showarticletitle{Joint face detection and alignment using multitask
  cascaded convolutional networks}.
\newblock \bibinfo{journal}{\emph{IEEE Signal Processing Letters}}
  \bibinfo{volume}{23}, \bibinfo{number}{10} (\bibinfo{year}{2016}),
  \bibinfo{pages}{1499--1503}.
\newblock


\bibitem[Zhang et~al\mbox{.}(2018)]%
        {zhang2018unreasonable}
\bibfield{author}{\bibinfo{person}{Richard Zhang}, \bibinfo{person}{Phillip
  Isola}, \bibinfo{person}{Alexei~A Efros}, \bibinfo{person}{Eli Shechtman},
  {and} \bibinfo{person}{Oliver Wang}.} \bibinfo{year}{2018}\natexlab{}.
\newblock \showarticletitle{The unreasonable effectiveness of deep features as
  a perceptual metric}. In \bibinfo{booktitle}{\emph{Proceedings of the IEEE
  conference on computer vision and pattern recognition}}.
  \bibinfo{pages}{586--595}.
\newblock


\bibitem[Zhao et~al\mbox{.}(2018)]%
        {zhao2018identity}
\bibfield{author}{\bibinfo{person}{Yajie Zhao}, \bibinfo{person}{Weikai Chen},
  \bibinfo{person}{Jun Xing}, \bibinfo{person}{Xiaoming Li},
  \bibinfo{person}{Zach Bessinger}, \bibinfo{person}{Fuchang Liu},
  \bibinfo{person}{Wangmeng Zuo}, {and} \bibinfo{person}{Ruigang Yang}.}
  \bibinfo{year}{2018}\natexlab{}.
\newblock \showarticletitle{Identity preserving face completion for large
  ocular region occlusion}.
\newblock \bibinfo{journal}{\emph{arXiv preprint arXiv:1807.08772}}
  (\bibinfo{year}{2018}).
\newblock


\bibitem[Zou et~al\mbox{.}(2021)]%
        {zou2021progressive}
\bibfield{author}{\bibinfo{person}{Xueyan Zou}, \bibinfo{person}{Linjie Yang},
  \bibinfo{person}{Ding Liu}, {and} \bibinfo{person}{Yong~Jae Lee}.}
  \bibinfo{year}{2021}\natexlab{}.
\newblock \showarticletitle{Progressive temporal feature alignment network for
  video inpainting}. In \bibinfo{booktitle}{\emph{Proceedings of the IEEE/CVF
  Conference on Computer Vision and Pattern Recognition}}.
  \bibinfo{pages}{16448--16457}.
\newblock


\end{thebibliography}

\appendix

\end{document}